\documentclass{article}

\usepackage{microtype}
\usepackage{graphicx}
\usepackage{subfigure}
\usepackage{url}
\usepackage{booktabs} 

\usepackage{hyperref}

\usepackage[accepted]{icml2023}

\usepackage{amsmath}
\usepackage{amssymb}
\usepackage{mathtools}
\usepackage{amsthm}

\usepackage[capitalize,noabbrev]{cleveref}

\theoremstyle{plain}

\theoremstyle{definition}

\theoremstyle{remark}

\usepackage[textsize=tiny]{todonotes}

\icmltitlerunning{Mist: Towards Improved Adversarial Examples for Diffusion Models}

\begin{document}

\twocolumn[
\icmltitle{Mist: Towards Improved Adversarial Examples for Diffusion Models}



\icmlsetsymbol{equal}{*}

\begin{icmlauthorlist}
\icmlauthor{Chumeng Liang}{equal,xxx}
\icmlauthor{Xiaoyu Wu}{equal,yyy,xxx}

\end{icmlauthorlist}

\icmlaffiliation{xxx}{cheer4creativity.ai}
\icmlaffiliation{yyy}{Shanghai Jiao Tong University, China}

\icmlcorrespondingauthor{Chumeng Liang}{caradryan2022@gmail.com}
\icmlcorrespondingauthor{Xiaoyu Wu}{wuxiaoyu2000@sjtu.edu.cn}

\icmlkeywords{Machine Learning, ICML}

\vskip 0.3in
]



\printAffiliationsAndNotice{\icmlEqualContribution} 

\begin{abstract}

Diffusion Models (DMs) have empowered great success in artificial-intelligence-generated content, especially in artwork creation, yet raising new concerns in intellectual properties and copyright. For example, infringers can make profits by imitating non-authorized human-created paintings with DMs. Recent researches suggest that various adversarial examples for diffusion models can be effective tools against these copyright infringements. However, current adversarial examples show weakness in transferability over different painting-imitating methods and robustness under straightforward adversarial defense, for example, noise purification. We surprisingly find that the transferability of adversarial examples can be significantly enhanced by exploiting a fused and modified adversarial loss term under consistent parameters. In this work, we comprehensively evaluate the cross-method transferability of adversarial examples. The experimental observation shows that our method generates more transferable adversarial examples with even stronger robustness against the simple adversarial defense. 


\end{abstract}

\section{Introduction}

Diffusion models (DMs)~\cite{sohl2015deep,ho2020denoising,song2020score} have demonstrated their great superiority in image synthesis~\cite{dhariwal2021diffusion}, especially in creating high-quality artwork~\cite{rombach2022high}. The success of DMs yields a boost in the field of digital art yet raises concerns about the copyright of human-created artwork. Since DMs offer convenient tools for artwork imitation and art style transfer~\cite{gal2022image,ruiz2022dreambooth}, infringers can make profits from generating artwork based on unauthorized human-created artwork. 

Adversarial examples for DMs~\cite{liang2023adversarial,shan2023glaze,van2023anti} are then born to prevent these malicious~\textit{scenarios} of images with DMs. By adding subtle perturbation to images, images are transferred into adversarial examples and not able to be learned and imitated by DMs. However, existing adversarial examples are tool-specific and not transferable over different scenarios, respectively. For example, \cite{salman2023raising} works well in image-to-image generation~\cite{rombach2022high} but fails in textual inversion~\cite{gal2022image}. 

In this paper, we investigate the cross-scenario performance of adversarial examples for diffusion models. We surprisingly find that a weighted combination of two adversarial examples~\cite{liang2023adversarial} attain strong transferability over three main scenarios of DM-based image imitation: Dreambooth~\cite{ruiz2022dreambooth}, textual inversion~\cite{gal2022image}, and image-to-image~\cite{rombach2022high}. We also propose that the performance of targeted adversarial examples is sensitive to the choice of targeted images. We further conduct experiments to investigate how these settings of adversarial examples impact the transferable performance and robustness and conclude a benchmark for selecting hyperparameters and targeted images. Based on these findings, we have open-sourced a pipeline to generate our state-of-the-art adversarial example as an online application, Mist. Mist is currently available on GitHub: \url{https://github.com/mist-project/mist}.

\section{Methods}

In this section, we mainly introduce two tricks for improving adversarial examples for diffusion models. The first trick introduces an effective approach to combine two terms of existing adversarial loss. The second focuses on picking a compatible target image for generating targeted adversarial examples.

\subsection{Combining Two Adversarial Losses}

In this part, we re-formulate two adversarial loss terms~\cite{liang2023adversarial,salman2023raising}. We explore combining these two loss terms as our optimization objective and exploiting the objective to generate more powerful adversarial examples for DMs. 

\subsubsection{Semantic Loss}

AdvDM~\cite{liang2023adversarial} proposes to exploit the training loss of diffusion models as the loss in generating adversarial examples. Concretely, it maximizes the training loss under certain sampling of latent variable $x'_{1:T}$ by fine-tuning the input $x$. 

\begin{equation}
\label{eq:semantic}
\begin{aligned}
    \delta &:= \arg\min\limits_{\delta} \mathbb{E}_{x'_{1:T}\sim u(x'_{1:T})}\mathcal{L}_{DM}(x',\theta),\\
    &\mbox{where } x\sim q(x), x'=x+\delta.
\end{aligned}
\end{equation}

Since $\mathcal{L}_{DM}(x',\theta)=\mathbb{E}_{t,\epsilon\sim \mathcal{N}(0,1)}[\Vert\epsilon - \epsilon_{\theta}(x'_t, t)\Vert^2_2]$, we exchange two expectation terms empirically and conclude the exact loss term as:

\begin{equation}
\label{eq:loss}
\begin{aligned}
     \mathbb{E}_{t,\epsilon\sim \mathcal{N}(0,1)}\mathbb{E}_{x'_{t}\sim u(x'_{t})}[\Vert\epsilon - \epsilon_{\theta}(x'_t, t)\Vert^2_2]\\
\end{aligned}
\end{equation}

where the expectation is estimated by Monte Carlo. Intuitively, this loss tries to pull the representation of the image $x$ out of the semantic space of the diffusion model. Our empirical observation indicates that the maximization of this loss results in chaotic content in the generated image based on adversarial examples. For this reason, we denote this loss as the \textit{semantic loss}.

\subsubsection{Textual Loss}

Another term of loss being widely discussed focuses on the encoding layer widely used in latent diffusion models (LDMs). LDM is a state-of-the-art variance of DMs that exploits an encoder and a decoder to map the image to representation in a latent space, where the diffusion process is then conducted. By reducing the dimension of the latent space, LDM significantly lowers the cost of both training and inference.

\begin{figure*}[htbp]

\begin{center}
\includegraphics[width=0.90\textwidth]{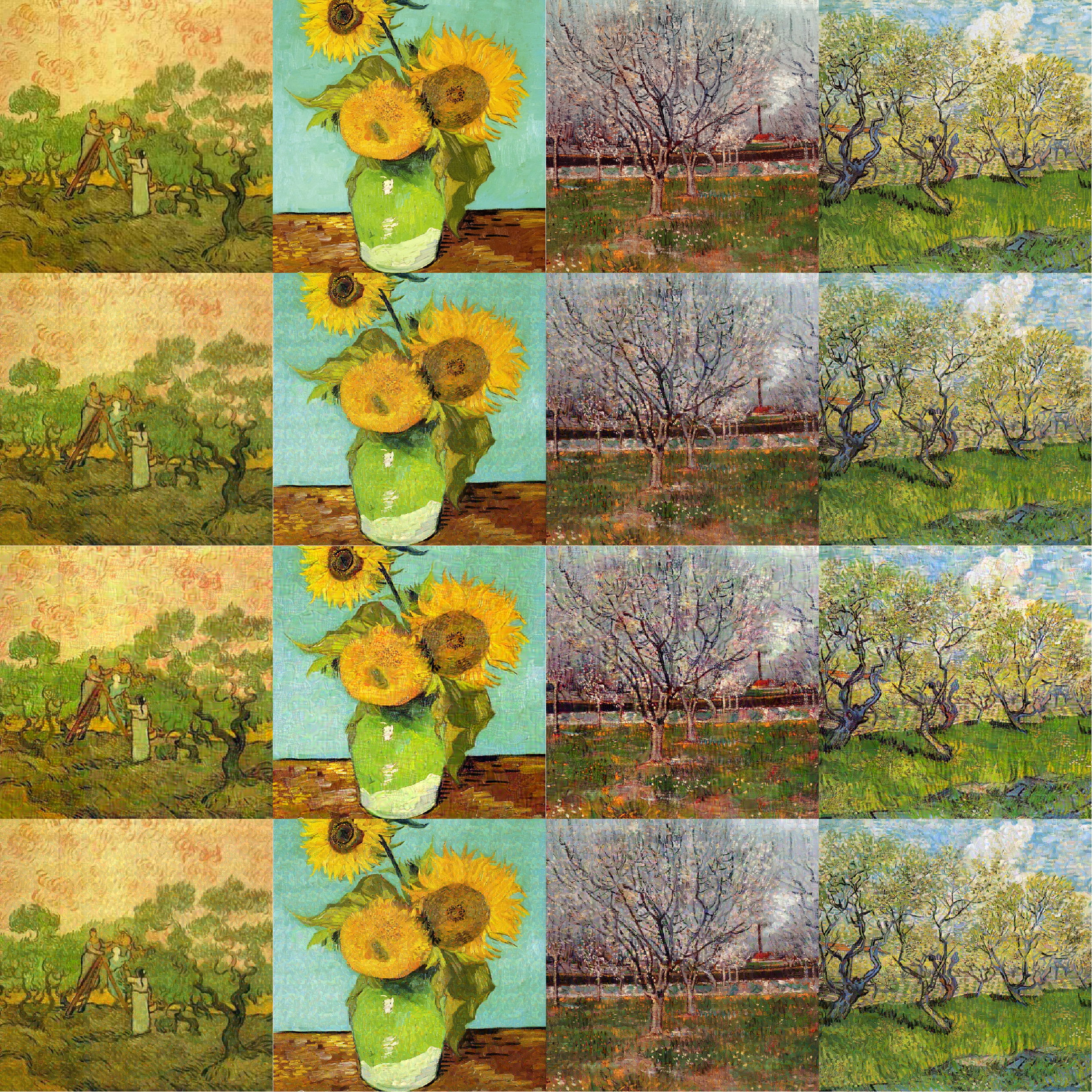}
\caption{\textbf{The first row:} Clean examples of Vincent Van Gogh's paintings. \textbf{The second row:} Adversarial examples of Vincent Van Gogh's paintings generated using the textural mode. \textbf{The third row:} Adversarial examples of Vincent Van Gogh's paintings generated using the semantic mode. \textbf{The fourth row:} Adversarial examples of Vincent Van Gogh's paintings generated using the fused mode with the fused weight set to $10^{4}$.}

\label{vangogh-sample}
\end{center}
\vskip -0.5in
\end{figure*}

This encoding layer provides an end-to-end process for the generation of adversarial examples~\cite{liang2023adversarial,}. Specifically, an adversarial example can be generated by adding subtle perturbation to maximize the distance between the encoded representation of the original image and that of the perturbed image.

    \begin{equation}
    \label{eq:textual}
       \begin{aligned}
       \delta:=&\arg\min\limits_{\delta}\mathcal{L}_\mathcal{E}(x, \delta, y)\\ =& \arg\min\limits_{\delta}\Vert\mathcal{E}(y) - \mathcal{E}(x + \delta)\Vert_2,
       \end{aligned}
    \end{equation}

where $\mathcal{E}$ denotes the image encoder of the latent diffusion model, $x$ represents the input image, and $y$ is the given target image. To optimize this loss, we employ the Projected Gradient Descent (PGD) attack~\cite{pgd}. The resulting perturbation exhibits characteristics resembling an embedded watermark on the background (refer to Figure \ref{vangogh-sample}). Hence, we denote this loss as the \textit{textural} loss.

\subsubsection{Joint Loss}

Both semantic loss and textual loss discussed provide unique advantages. In light of this, we explore combining these targets to create a new objective loss function. We merge the two targets to form the following objective loss:

\begin{equation}
\label{eq:fused}
\begin{aligned}
        \delta:= \arg\max\limits_{\delta}&(w\mathbb{E}_{x'_{1:T}\sim u(x'_{1:T})} \mathcal{L}_{DM}(x',\theta) - \mathcal{L}_\mathcal{E}(x, \delta, y)) ,\\
    &\mbox{where } x\sim q(x), x'=x+\delta.
 \end{aligned}
\end{equation}

where $w$ represents the fused rate. We have known that the semantic loss $\mathcal{L}_{DM}(x',\theta)$ consists of an expectation term estimated by Monte Carlo. The main problem to optimize the combined loss term is determining how to jointly optimize the semantic loss and the textual loss. We find that computing textual loss every time the semantic loss is estimated on the sampled $t$ works well empirically. The final loss term used in Mist can be concluded as follows: 

\begin{equation}
\label{eq:final}
\begin{aligned}
   \mathbb{E}_{t,\epsilon\sim \mathcal{N}(0,1)}&\mathbb{E}_{x'_{t}\sim u(x'_{t})}
   [w\Vert\epsilon - \epsilon_{\theta}(x'_t, t)\Vert^2_2\\
   -&\Vert\mathcal{E}(y) - \mathcal{E}(x + \delta)\Vert_2]\\     
 \end{aligned}
\end{equation}

In the implementation of Mist, we provide three modes, corresponding to three terms of adversarial loss. Semantic and textual mode corresponds to semantic and textual loss, respectively. Fused mode corresponds to the combined semantic and textual loss.

\subsection{Selecting targeted images in Textual Loss is critical for robustness and transferability}

As shown in Eq~(\ref{eq:textual}), we include the targeted image $y$ as a variable in the textual loss. Our observation shows that the performance of textual loss is impressively sensitive to the targeted image (See in Sec~\ref{sec:diff_tar}). Empirically, it is better to select images with high contrast ratio and sharp canny as the targeted image $y$. We visualize the effect comparison of different choices of the targeted image in Fig~\ref{target-compare}. Note that an appropriate choice of the targeted image can not only improve the effectiveness of adversarial examples but also its robustness against noise purification.

\section{Experiments}

In this section, we evaluate Mist, the proposed method for generating adversarial examples for Stable Diffusion Model~\footnote{https://github.com/CompVis/stable-diffusion}~\cite{rombach2022high}. We  use the $l_{\infty}$ norm as the constraint for generating all the adversarial examples. Following existing research in adversarial examples, we set the sampling step as 100, the per-step perturbation budget as 1/255 and the total budget as 17/255. Our experiments mainly use  Van Gogh's paintings collected from WikiArt~\cite{wikiart}. The default mode for Mist is the fused mode, with a default fused weight of  1e4. All experiments were conducted using an NVIDIA RTX 3090 GPU. We then evaluate the effects of Mist in various scenarios, including pre-training cases like textual inversion~\cite{gal2022image}, dreambooth~\cite{ruiz2022dreambooth}, and Scenario.gg~\footnote{https://app.scenario.gg/}, where Mist serves as a protective approach against image style transfer. Additionally, we assess its performance in preventing image modifications from image-to-image applications like NovelAI~\footnote{https://novelai.net/}.

\subsection{Effects of Mist under different scenarios}

\begin{figure*}[htbp]
\vskip 0.2in
\begin{center}
\includegraphics[width=1.00\textwidth]{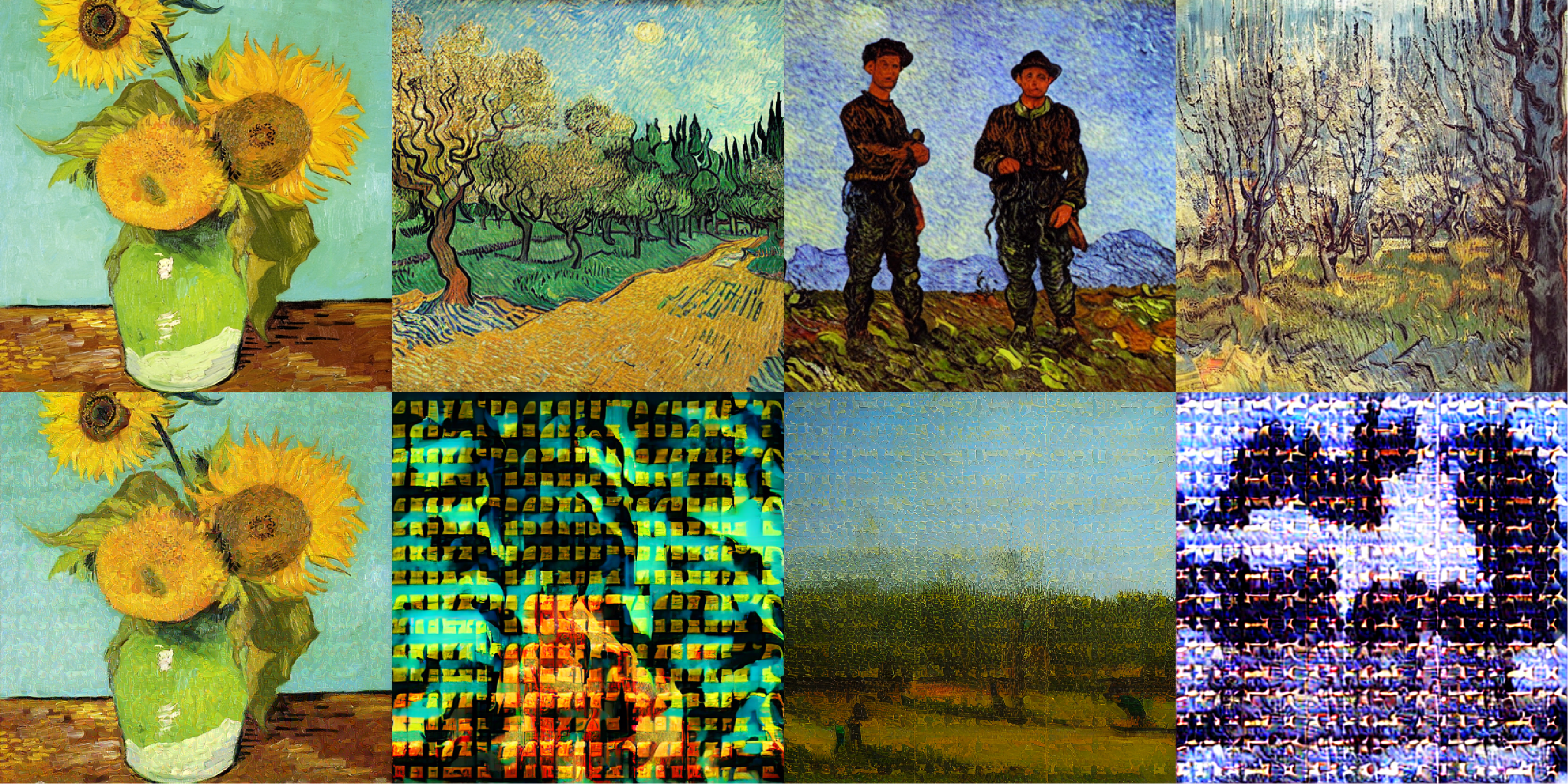}
\caption{Effects of Mist under pre-trained scenarios. \textbf{From left to right:} Source images, generated images under textual inversion, generated images under dreambooth, generated images under scenario.gg. \textbf{The first row:} Source and generated images for Van Gogh's paintings. \textbf{The second row:} Source and generated images for attacked Van Gogh's paintings. 
}
\label{effectiveness}
\end{center}
\vskip -0.2in
\end{figure*}

\begin{figure*}[htbp]
\vskip 0.2in
\begin{center}
\includegraphics[width=1.00\textwidth]{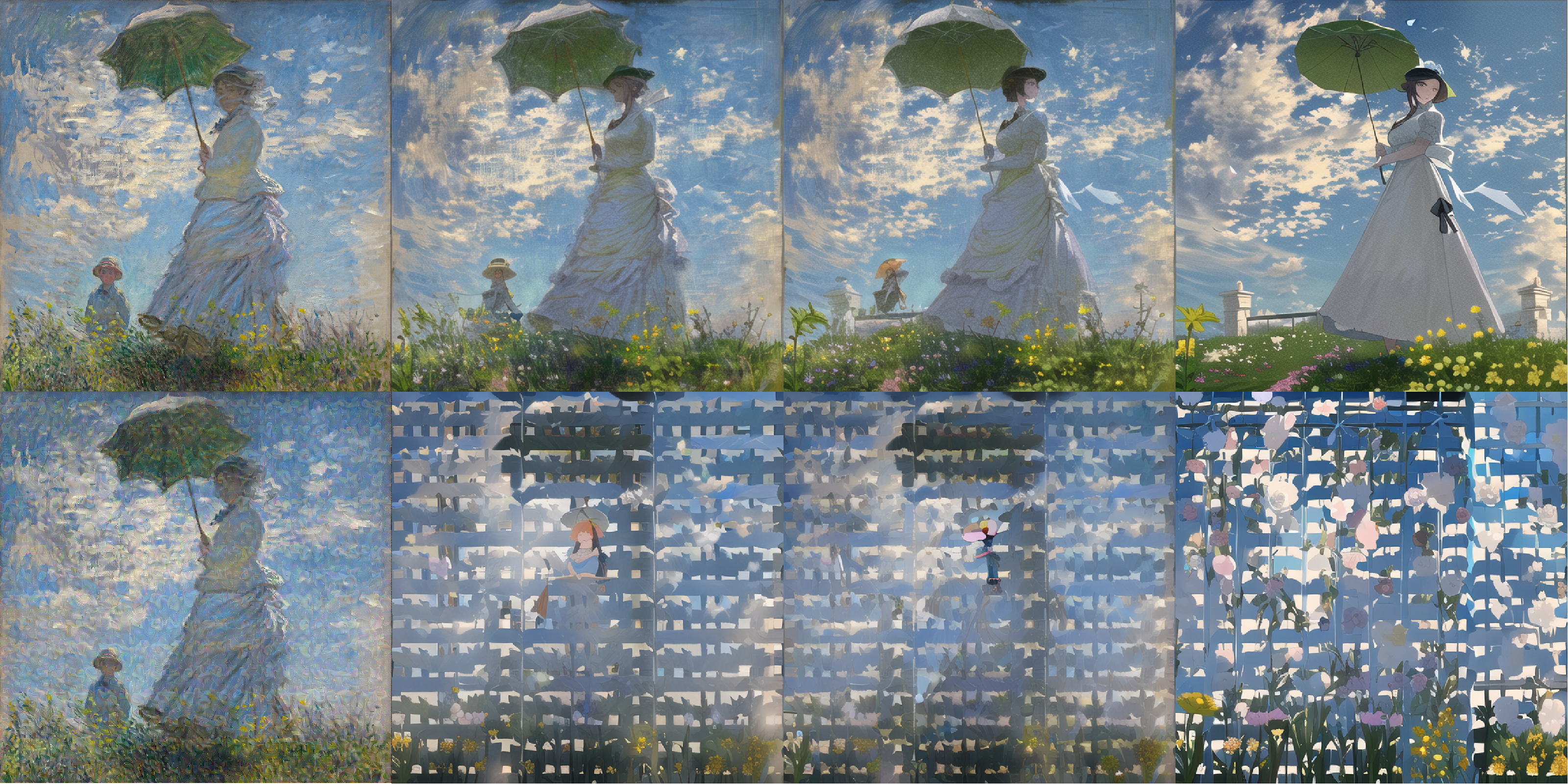}
\caption{Effects of Mist under NovelAI image-to-image. \textbf{From left to right:} Source images, generated images with strength 0.25, generated images with strength 0.35, generated images with strength 0.5. \textbf{The first row:} Source and generated images for Monet's paintings. \textbf{The second row:} Source and generated images for attacked Monet paintings. }
\label{effectivenessi2i}
\end{center}
\vskip -0.2in
\end{figure*}

\begin{figure*}[htbp]
\vskip 0.2in
\begin{center}
\includegraphics[width=1.00\textwidth]{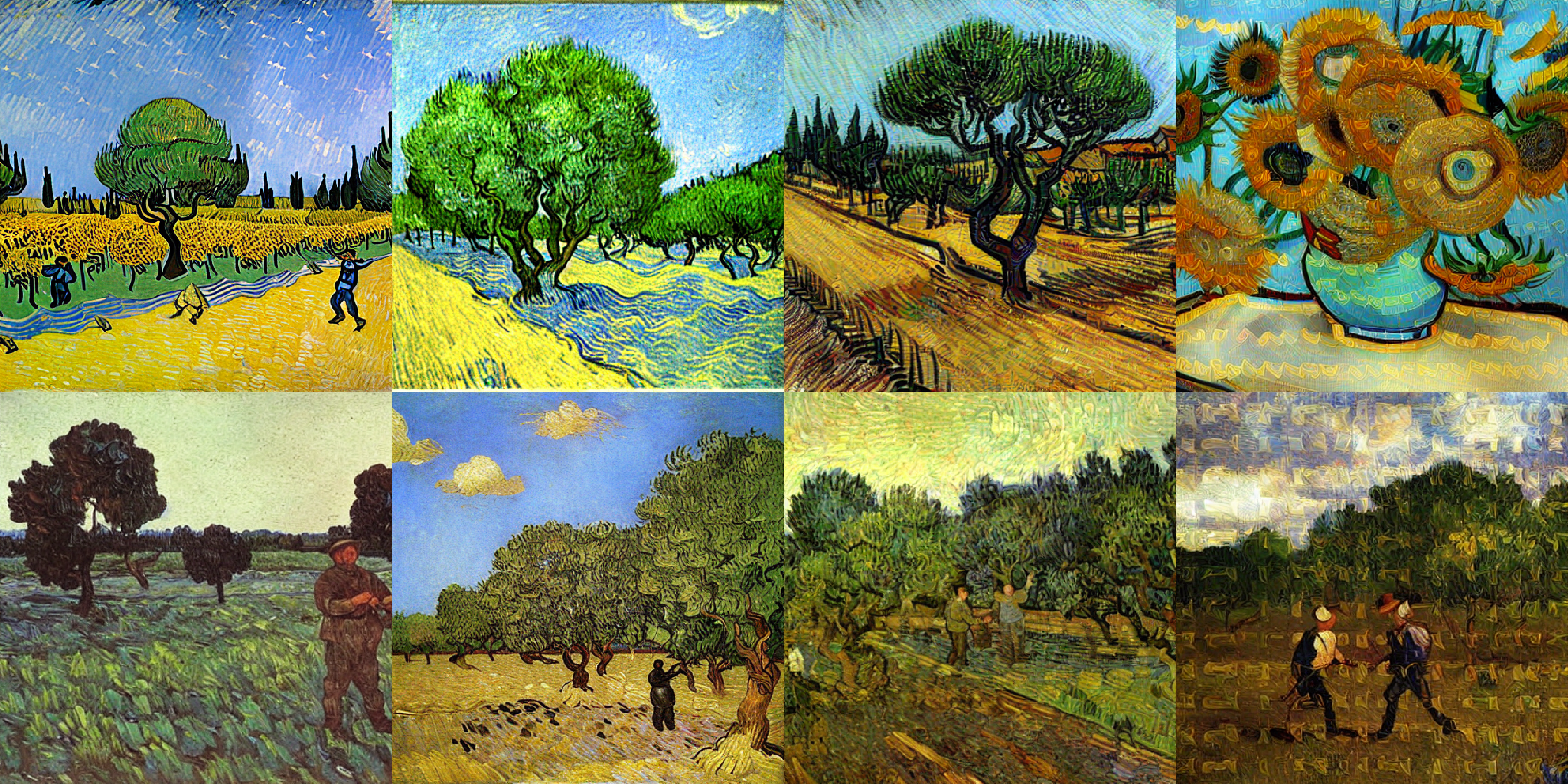}
\caption{Comparison of the robustness of different methods. \textbf{From left to right:} Generated images based on clean images, gaussian-perturbed images, glazed-images and Misted images. \textbf{The first row:} Generate images based on clean and adversarial examples under textual inversion. \textbf{The second row:}Generate images based on clean and adversarial examples under dreambooth. }

\label{robustness}
\end{center}
\vskip -0.2in
\end{figure*}
We conduct qualitative experiments to evaluate the effects of Mist under pre-trained scenarios. In the  textural inversion scenario, we fix the number of vectors per token to 8 and train the embedding for 6,000 steps using the given images. For dreambooth, we retrain both the UNet and the text encoder in  Stable diffusion v1.4 . The learning rate is fixed at 2e-6, and the maximum training steps are set to 2000. In the experiments involving the style transferring tool from scenario.gg,  we utilized the auto training mode and selected Art style - Painting as the training class.  More details can be found in our documentation~\footnote{https://mist-documentation.readthedocs.io/en/latest}." As is shown in Figure \ref{effectiveness}, Mist effectively protects images from AI-for-Art-based mimicry.

We also conducted experiments under the NovelAI image-to-image scenario. We utilized the NAI Diffusion Anime model and set the prompt to 'woman with a Parasol, high resolution, outdoor, flowers, blue sky'.  We set the resolution to 512, the random seed to 1255, the steps to 40, the guidance to 11, and the sampler to DPM++ 2M. Then we change the strength to 0.25, 0.35 and 0.5 respectively. From Figure \ref{effectivenessi2i}, we can observe that Mist is also effective under image-to-image scenario.

Under pre-training scenarios, infringers can generate high-quality images with cropped-and-resized images. However, such preprocessing process may destroy the semantic information carried by adversarial perturbations and disable the attack. Based on this, we also conduct experiments on the robustness of Mist under preprocessing. For each images, we first crop 64 pixels in all directions and resize the images back into 512 $\times$ 512 resolution. We compare the robustness of our method under such input transformation with gaussian noise and glaze~\footnote{https://glaze.cs.uchicago.edu/}~\cite{shan2023glaze} under textual inversion and dreambooth. Both the  gaussian noise and Mist are constrained with $17/255$ budget in $L_{\infty} $ norm. Glaze(with very high intensity and medium render quality) is constrained with $20/255$ budget in $L_{\infty} $ norm . From Figure \ref{robustness}, we can observe that Mist is the only method remains effective under crop-and-resize input transformation.

\subsection{Comparison of Mist under different modes}

It is interesting that which mode of Mist is more effective under different scenario. As stated before, the intuition of our fused mode is to make Mist applicable under all scenarios.

Towards this end, we conduct experiments to compare different modes of Mist. We follow the experiment setting mentioned in previous section and generate 50 images for each embedding (under textual inversion) or model weight checkpoint (under dreambooth). Then we evaluate the sample quality of generated images by two metrics: Fréchet Inception Distance (FID) and Precision ($prec.$). The FID and $prec.$ between the generated images and the source images are computed to compare the strength of different modes of Mist.

\begin{figure*}[htbp]

\begin{center}
\includegraphics[width=0.90\textwidth]{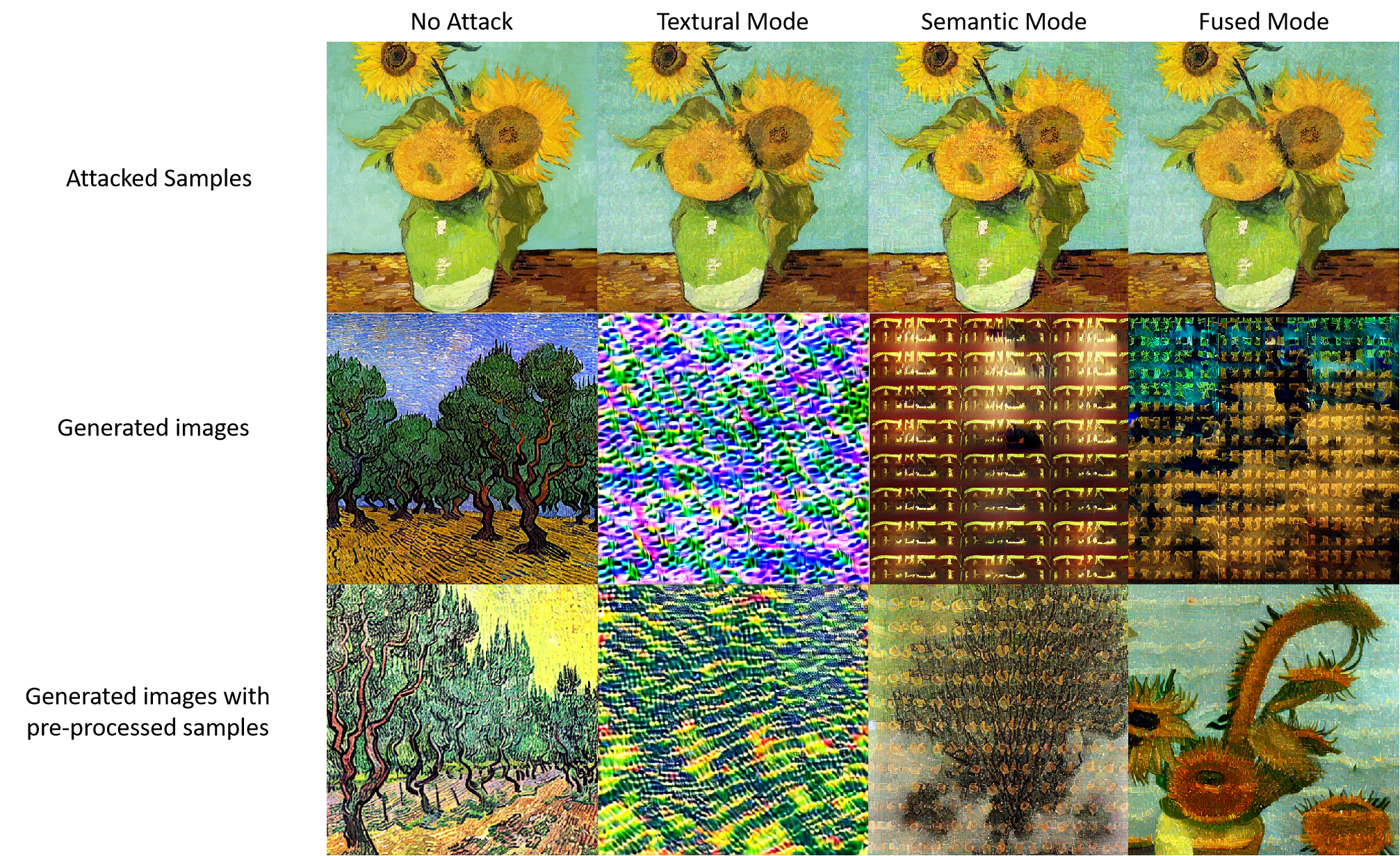}
\caption{Comparison of different modes of Mist under textual inversion. The fused weight $w$ for the fused mode is set to $10^{4}$.\textbf{The first row:} Adversarial examples of Van Gogh's paintings under different modes \textbf{The second row:} Generated images based on attacked Van Gogh's paintings. \textbf{The third row:}Generated images based on pre-processed attacked Van Gogh's paintings. }

\label{ti-gen}
\end{center}
\vskip -0.5in
\end{figure*}

From Table \ref{tab:mode-com-ti} and Figure \ref{ti-gen}, it is evident that the semantic mode demonstrates the highest effectiveness and robustness under the textual inversion scenario. The semantic mode is expected to be one of the most potent attacks when the model weight $\theta$ remains unchanged. Further details on this can be found in our previous work \cite{liang2023adversarial}. Textual inversion specifically relies on an embedding that handles text-modal information and does not alter the weight of the model's backbone. This characteristic makes the semantic mode particularly well-suited for the textual inversion scenario.

\begin{figure*}[htbp]

\begin{center}
\includegraphics[width=0.90\textwidth]{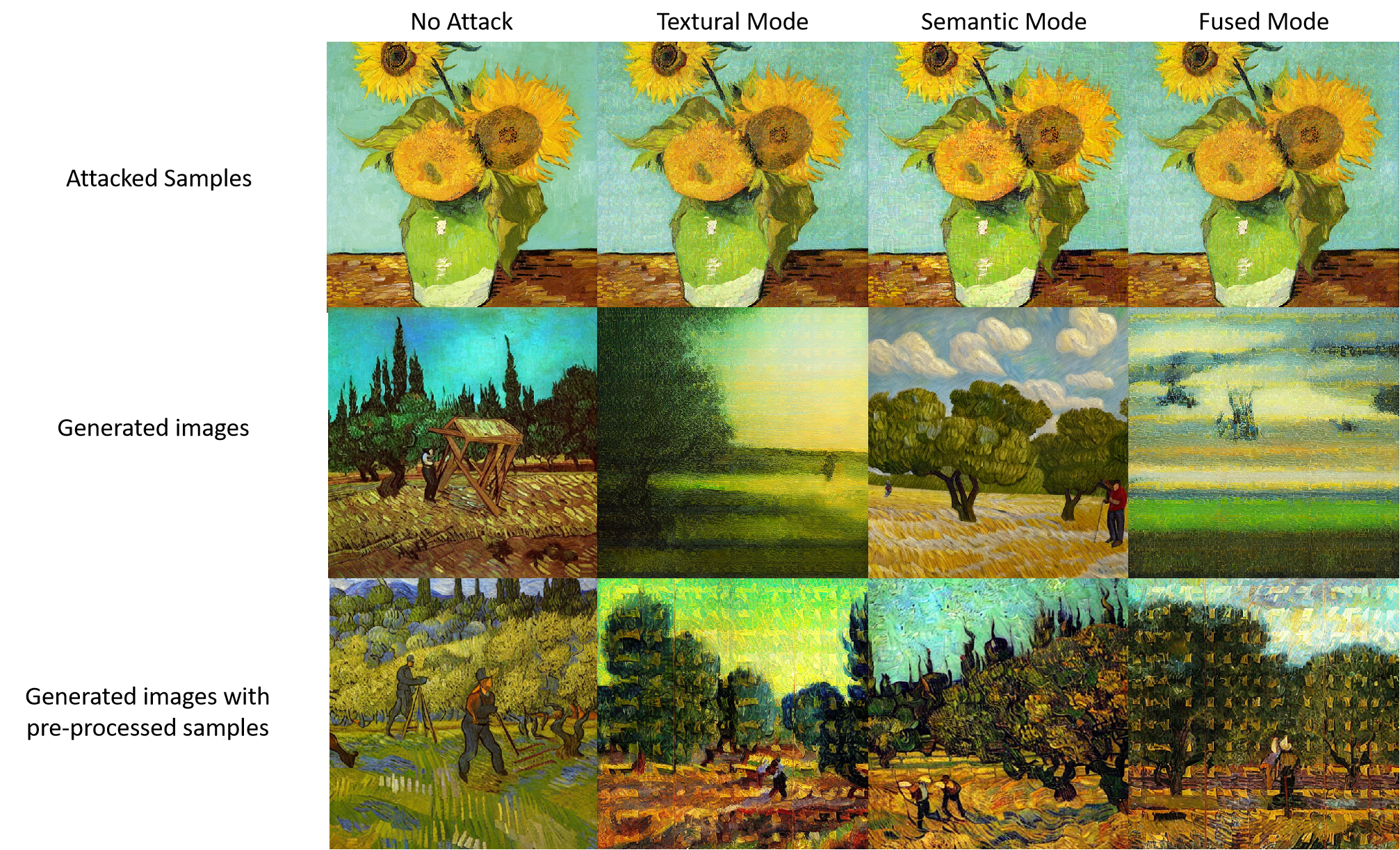}
\caption{Comparison of different modes of Mist under dreambooth. The fused weight $w$ for the fused mode is set to $10^{4}$. \textbf{The first row:} Adversarial examples of Van Gogh's paintings under different modes \textbf{The second row:} Generated images based on attacked Van Gogh's paintings. \textbf{The third row:}Generated images based on pre-processed attacked Van Gogh's paintings. }

\label{db-gen}
\end{center}
\vskip -0.5in
\end{figure*}

Under the dreambooth scenario, where the weight of the model backbone is changed, the semantic mode exhibits reduced effectiveness. As shown in Table \ref{tab:mode-com-db} and Figure \ref{db-gen}, the textual mode proves to be more effective and robust compared to the semantic mode. This difference in effectiveness can be attributed to two key factors:

\begin{enumerate}
    \item The image encoder employed in dreambooth significantly reduces the resolution of input images. This process is highly semantic and can be exploited for adversarial attacks.
    \item Most pre-trained methods, such as those mentioned in \cite{ruiz2022dreambooth}, do not modify the image encoder of the latent diffusion model. This is likely because the stable diffusion model trains the diffusion model component using the latent space of a frozen auto-encoder. Retraining the auto-encoder could potentially alter the latent space and degrade performance.
\end{enumerate}

\begin{table}[t]
\caption{Text-to-image generation based on textual inversion. Different modes are compared to generate adversarial examples. $w$ refers to the fused rate mentioned in Equation \ref{eq:fused}.}
\vspace{0.1in}
\label{tab:mode-com-ti}
\resizebox{1\linewidth}{!}{%
\begin{tabular}{ccccc}
\hline
\multicolumn{1}{l}{} & \multicolumn{2}{c}{No Preprocesing} & \multicolumn{2}{c}{Crop and Resize} \\
                     & FID$\uparrow$   & $prec.\downarrow$ & FID$\uparrow$   & $prec.\downarrow$ \\ \hline
No Attack            & 237.56          & 0.9               & 280.63          & 1                 \\
Textural             & 419.43          & 0.04              & 303.05          & 0.72              \\
Fused($w=10^3$)      & 454.39          & 0.02              & 297.90          & 0.80              \\
Fused($w=10^4$)      & 371.12          & 0.22              & 277.88          & 0.44              \\
Fused($w=10^5$)      & 416.25          & 0.04              & 320.52          & 0.70              \\
Semantic             & \textbf{465.82} & \textbf{0.02}     & \textbf{350.87} & \textbf{0.18}     \\ \hline
\end{tabular}%
}

\end{table}

The fused mode of Mist combines the textural and semantic modes, resulting in a balanced performance under textual inversion and dreambooth scenarios. The choice of the fused weight parameter, denoted as $w$, plays a crucial role in determining the performance of the fused mode.

In general, a higher fused weight $w$ leads to performance similar to the semantic mode, with higher effectiveness under textual inversion and lower effectiveness under dreambooth. Conversely, a lower fused weight $w$ brings the performance closer to the textural mode, with higher effectiveness under dreambooth and lower effectiveness under textual inversion.

However, it should be noted that the performance of the fused mode does not strictly follow a consistent pattern with changes in the fused weight $w$. This could be attributed to the fact that the two different targets of the textural and semantic modes may not be entirely consistent and might partially interfere with each other. To gain a better understanding of this phenomenon, further detailed experiments are required to validate these hypotheses.

\begin{table}[t]
\caption{Text-to-image generation based on dreambooth. Different modes are compared to generate adversarial examples.$w$ refers to the fused rate mentioned in Equation \ref{eq:fused}.}
\label{tab:mode-com-db}
\vspace{0.1in}
\resizebox{1\linewidth}{!}{%
\begin{tabular}{ccccc}
\hline
\multicolumn{1}{l}{} & \multicolumn{2}{c}{No Preprocesing} & \multicolumn{2}{c}{Crop and Resize} \\
                     & FID$\uparrow$   & $prec.\downarrow$ & FID$\uparrow$   & $prec.\downarrow$ \\ \hline
No Attack            & 274.40     & 0.88              & 279.54     & 0.88              \\
Textural             & 392.94          & 0.26              & \textbf{353.86} & 0.58              \\
Fused($w=10^3$)      & 429.40          & \textbf{0.04}     & 347.10          & 0.52              \\
Fused($w=10^4$)      & \textbf{444.92} & 0.10              & 340.20          & 0.72              \\
Fused($w=10^5$)      & 357.08          & 0.26              & 328.44          & \textbf{0.46}     \\
Semantic             & 376.15          & 0.38              & 267.30          & 0.96              \\ \hline
\end{tabular}%
}

\end{table}

\subsection{Comparison of different target images for textural mode}
\label{sec:diff_tar}

\begin{figure*}[t]
\vskip -4.5in
\begin{center}
\includegraphics[width=0.90\textwidth]{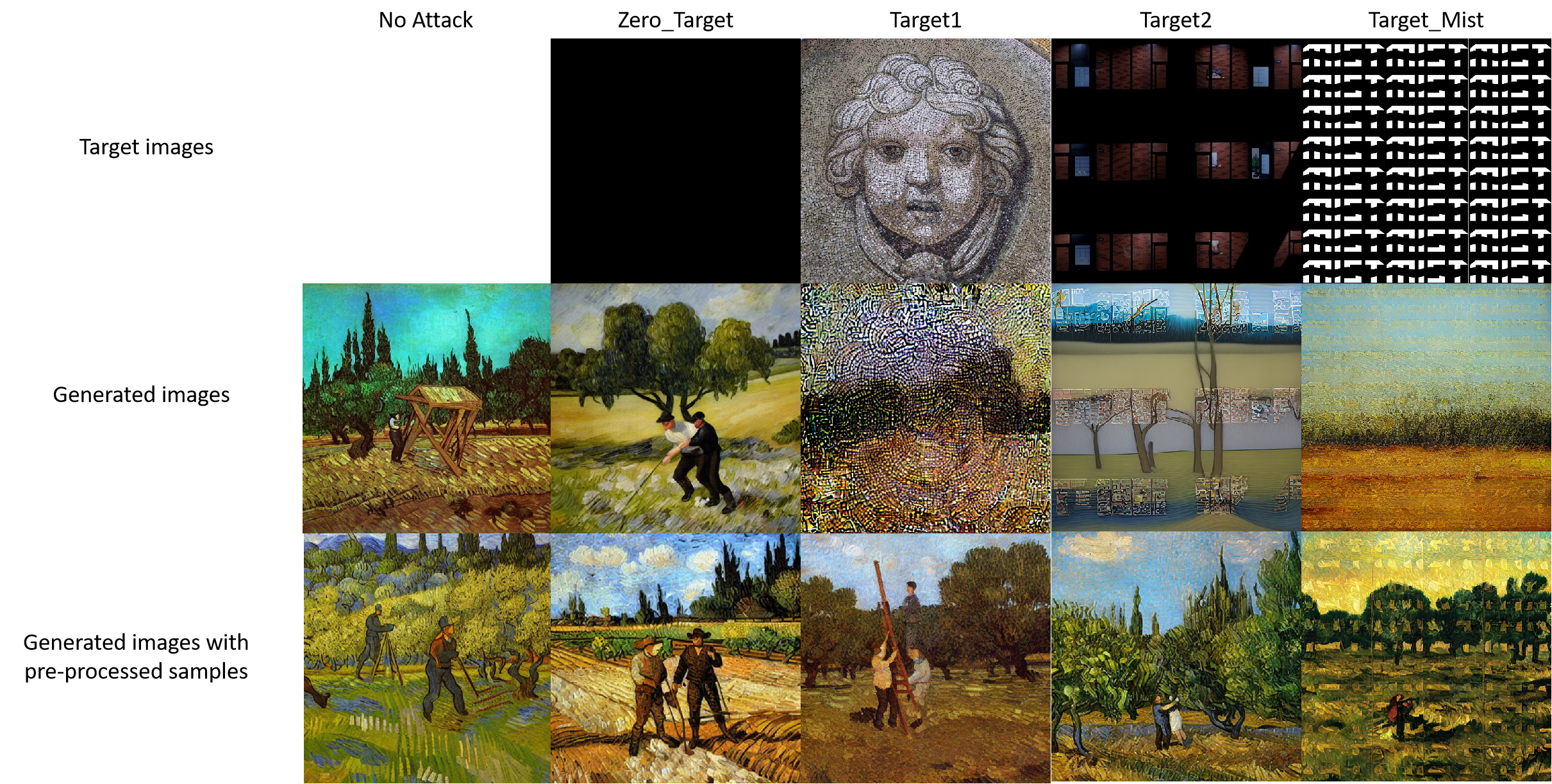}
\caption{\textbf{The first row:}The target images for Mist under textural m \textbf{The second row:} Generated images based on attacked Van Gogh's paintings. \textbf{The third row:}Generated images based on pre-processed attacked Van Gogh's paintings.  Among all the target images, only the densely arranged pattern of "MIST" remains effective under pre-processing.}

\label{target-compare}
\end{center}
\vskip -0.05in
\end{figure*}

We also find that the choice of target images is closely related to the robustness of Mist under textural and fused mode. We choose four different target images: a black image with no information  (Zero\_Target), a photo of  sculpture Art at the Sistine Chapel in Rome (Target1), a photo of structural architecture (Target2) and the densely arranged pattern of "MIST" logo (Target\_Mist). We follow the settings in previous sections under dreambooth. From Figure \ref{target-compare} and Table 
\ref{tab:target-comparison}, we can observe the following:

\begin{table}[t]
\caption{Text-to-image generation based on dreambooth. Different target images are selected to generate adversarial examples using the textural mode.}
\vspace{0.1in}
\label{tab:target-comparison}
\resizebox{1\linewidth}{!}{%
\begin{tabular}{ccccc}
\hline
\multicolumn{1}{l}{} & \multicolumn{2}{c}{No Preprocesing}      & \multicolumn{2}{c}{Crop and Resize}      \\
                     & FID$\uparrow$        & $prec.\downarrow$ & FID$\uparrow$        & $prec.\downarrow$ \\ \hline
No Attack            & 274.40          & 0.88              & 279.54          & 0.88              \\
Zero\_Target         & 325.58           & 0.28              & 308.17          & 0.84              \\
Target1              & 380.31          & 0.02              & 336.47          & \textbf{0.44}     \\
Target2              & \textbf{497.54} & \textbf{0}        & 321.46         & 0.58              \\
Target\_Mist         & 392.93          & 0.26              & \textbf{353.85} & 0.58              \\ \hline
\end{tabular}%
}
\end{table}

\begin{enumerate}
    \item The Zero\_Target image is the least effective choice for the textual mode. This could be because the textual mode relies on implanting specific semantic information onto the latent space of input images to misguide the pre-training process of the diffusion model. Since the Zero\_Target image does not contain any specific semantic information, its performance is relatively low.
    \item Target images with high contrast (such as Target\_Mist and Target2) result in stronger attacks compared to those with low contrast (Zero\_Target and Target1).
    \item Images with repetitive patterns (Target1 and Target\_Mist) exhibit more robustness against input transformations.  This could be due to the specific frequency spectrum of these target images and further research is needed to fully understand this phenomenon.
\end{enumerate}

We only choose several representative figures as the target images. Larger-quantity experiments are needed for a deeper understanding of the textural mode of Mist.

\newpage
\bibliography{main}
\bibliographystyle{icml2023}

\newpage

\end{document}